\documentclass{article}

\usepackage[utf8]{inputenc}
\usepackage{amssymb,amsmath}
\usepackage{pifont}
\usepackage{graphicx,verbatimbox}
\usepackage{hyperref} 
\usepackage{tcolorbox}
\usepackage{makecell}

\usepackage{xspace}
\usepackage{booktabs}
\usepackage{multirow}
\usepackage{rotating}
\usepackage{color}
\usepackage{palatino}

\usepackage{caption} 
\newcommand{\cmark}{\ding{51}\xspace}%
\newcommand{\xmark}{\ding{55}\xspace}%

\title{ViT-P: Rethinking Data-efficient Vision Transformers from Locality}

\author{
\begin{minipage}{\linewidth}
\begin{center}
\large Bin Chen$^{\star}$ \hspace{0.2cm} Ran Wang$^\dagger$ \hspace{0.2cm} Di Ming$^{\star}$ \hspace{0.2cm} Xin Feng$^{\star}$\\[0.15cm] 
\scalebox{1.}{
    $^\star$Chongqing University of Technology\hspace{0.15cm}
    $^\dagger$University of California
}\\[0cm]
\end{center}
\end{minipage}
}

\date{~}

\begin{document}

\maketitle
\begin{abstract}
Recent advances of Transformers have brought new trust to computer vision tasks.
However, on small dataset, Transformers is hard to train and has lower performance than convolutional neural networks.
We make vision transformers as data-efficient as convolutional neural networks by introducing multi-focal attention bias.
Inspired by the attention distance in a well-trained ViT, we constrain the self-attention of ViT to have multi-scale localized receptive field.
The size of receptive field is adaptable during training so that optimal configuration can be learned.
We provide empirical evidence that proper constrain of receptive field can reduce the amount of training data for vision transformers.
On Cifar100, our ViT-P Base model achieves the state-of-the-art accuracy (83.16\%) trained from scratch. We also perform analysis on ImageNet to show our method does not lose accuracy on large data sets.
\end{abstract}

\section{Introduction}
\label{sec:introduction}

Following the deep convolutional neural network~\cite{NIPS2012_c399862d,he2016deep}, the vision transformer led a series of new breakthroughs in computer vision tasks~\cite{dosovitskiy2020image,touvron2021training,liu2021Swin,carion2020detr}. Vision transformer uses multi-head self-attention with global spatial attention to extract image features with less inductive bias~\cite{d2021convit}, and can continuously improve accuracy by increasing training data~\cite{dosovitskiy2020image}.
However, the real data set for visual tasks are often not big enough to fully exploit the capacity of vision transformers.
Recent evidence~\cite{liu2021Swin,d2021convit,zhang2021aggregating,dai2021coatnet} shows that it is important to introduce locality into vision transformers, which enables vision transformers to achieve comparable performance as convolutional neural networks with only 0.3\% of training data (from JFT~\cite{sun2017revisiting} of 300 million data to ImageNet~\cite{deng2009imagenet} of 1 million data).

On the other hand, Dosovitskiy et al.~\cite{dosovitskiy2020image} indicated that global architectures have more potential, although such solution can only be achieved with a large amount of training data.
One possibility of this appetite for data is that the global network does not learn local features but global features that are more generalized than local ones, and global features are harder to learn than local features.
But Cordonnier et al.~\cite{Cordonnier2020On} showed that vision transformers do perform convolution-like local attention in the shallow layers of the network, which negates the above point.
Additionally, this phenomenon of local attention was not observed on DeiT~\cite{touvron2021training} trained by Cifar~\cite{krizhevsky2009learning}.

\begin{figure}[t]
\centering
\includegraphics[width=1\linewidth]{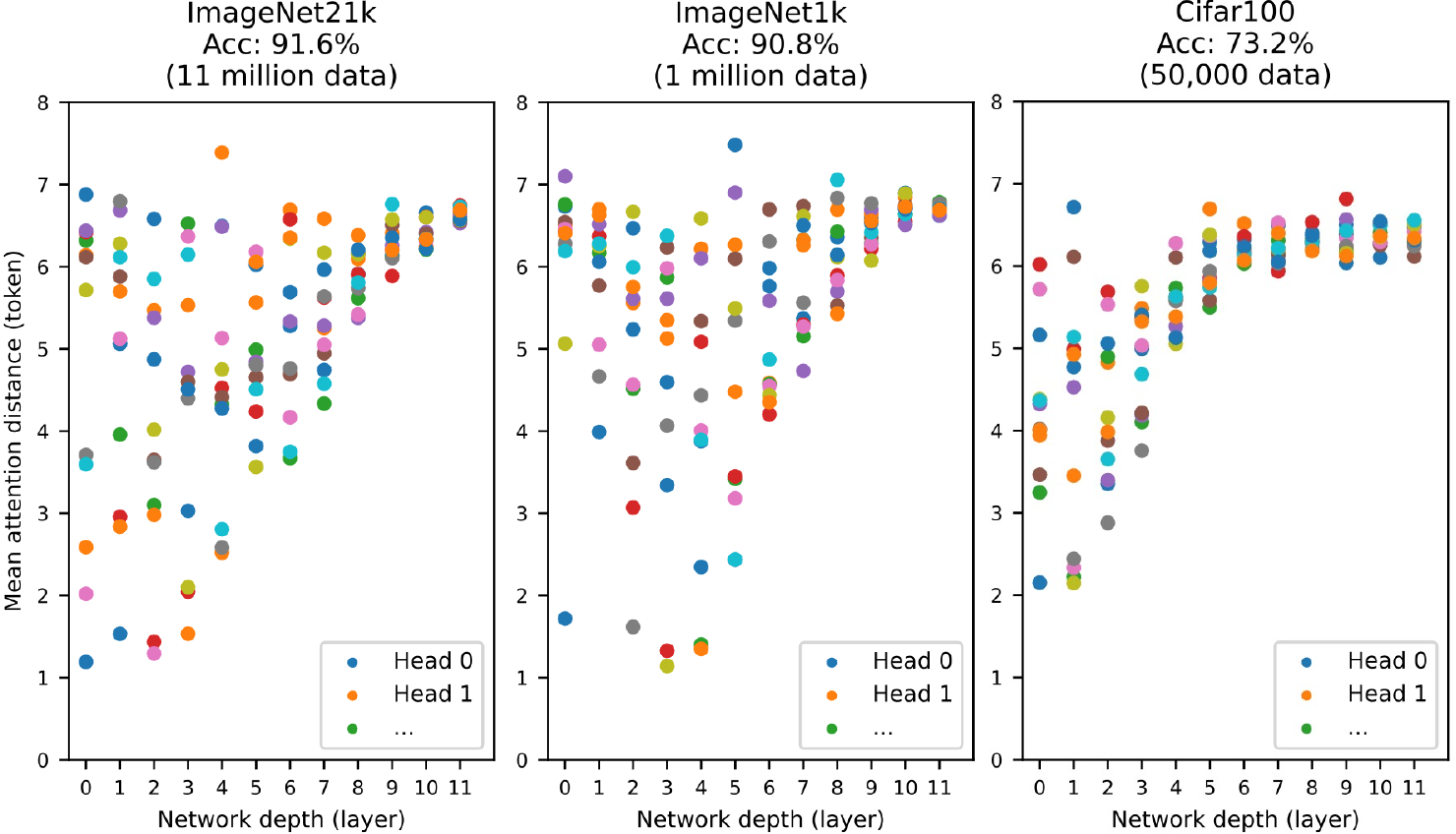}
\caption{
Size of attended area by head and network depth.
Each dot shows the mean attention distance (MAD)~\cite{dosovitskiy2020image} across images for one of $12$ heads at one layer.
Attention distance was computed by averaging the distance between the query token and all other tokens, weighted by the attention weight.
This ``attention distance" is analogous to receptive field size in CNNs.
\textit{These three illustrations are ViT-B pre-trained on data sets of different sizes and then fine-tuned on Cifar.}
Less training data has lower accuracy and fewer local attention heads.
Even with ImageNet1k (center) of 1 million data, the first two layers still lack local attention heads.
}
\label{fig:diff_train_data}
\end{figure}

To further explore this phenomenon, we analyze the receptive field of each attention head in transformers with mean attention distance (MAD) ~\cite{dosovitskiy2020image}.
When the network is pre-trained on data sets of different sizes and fine-tuned on the same Cifar data set, the problem of excessive long-range attention exposes: as the amount of data decreases, the accuracy gradually degrades to below convolutional neural networks, and the network begins to lack local attention heads. Unexpectedly, this lack of local attention heads is not due to under-fitting, and there is a consistent pattern for longer training.
Figure~\ref{fig:diff_train_data} shows a typical example.

The lack of local attention heads indicates that vision transformers require large amounts of data to learn attention heads with local features.
Consider an architecture, which can obtain the required local spatial attention head through initialization.
There exists a solution by introducing focal attention bias: the long-range attention of attention heads is suppressed, and attention heads have different receptive fields.
These different receptive fields can be extended in the depth or width of the network, which is similar to introducing multi-scale receptive fields for vision transformers in terms of implementation. Specifically, we discuss vision transformers with multi-scale receptive fields in terms of width, depth, and both.
There are works similar to these three modes. For example, ConViT~\cite{d2021convit} combines self-attention and convolution in width, Swin~\cite{liu2021Swin} gradually expands the receptive field with the down-sampling of the network, and ViT~\cite{dosovitskiy2020image} trained with large data sets has multi-scale receptive fields in width and depth.
There is no work comparing these three modes under the same framework. In Table~\ref{table:diff_mode}, we show the advantage of multi-scale receptive fields in width.

\begin{figure}[t]
\centering
\includegraphics[width=1\linewidth]{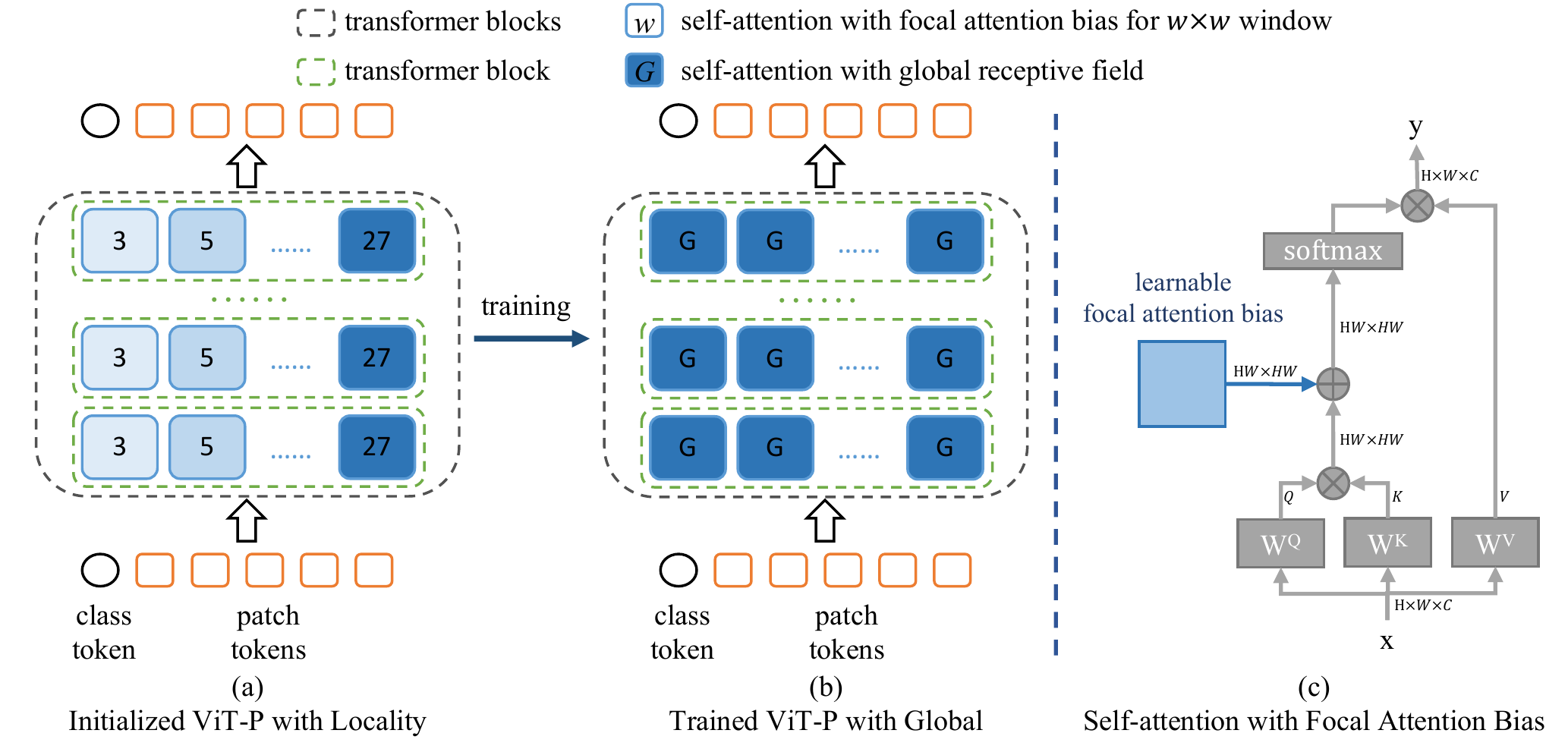}
\caption{
    Overview of ViT-P.
    At initialization, locality is introduced into the network by introducing a learnable focal attention bias for each self-attention head.
    And the receptive field configuration of self-attention heads is different  (e.g. MRFA-W).
    As training, the network adjusts the receptive field globally.
}
\label{fig:overview}
\end{figure}

Moreover, we consider the global network, which means that the global receptive field of the network is preserved while introducing locality.
Our approach is to make the focal attention bias learnable during training so that the receptive field of attention heads can expand or shrink.
We treat this network, which can globally adjust the receptive field, as a global solution. As shown in Figure~\ref{fig:overview}.

In summary, this paper assists the training (of the vision transformer with global receptive fields) by introducing locality to the self-attention module to improve the data efficiency of vision transformers.
We no longer hope the network to reduce excessive long-range attention just by learning through large amounts of data, but explicitly introduce the prior at initialization.
We hypothesize that self-attention introducing correct prior is easier to optimize than the original. To the extreme, if the local experience is optimal, it will be easier to suppress long-range attention directly than to fit through linear layers.

We evaluate the data efficiency of our approach on the popular small classification data set, Cifar~\cite{krizhevsky2009learning}, and compare it with the highly competitive Swin~\cite{liu2021Swin} and NesT~\cite{zhang2021aggregating}.
Swin and NesT have undergone many design iterations, and their performance has been greatly improved. Our experiments show that our new model achieves comparable performance.
More precisely, ViT-P performs better on small data sets, a result likely enabled by more efficient local inductive bias.
It is not outstanding compared to them on large data sets (such as ImageNet~\cite{deng2009imagenet}), but it does not prevent us from replacing convolutional neural networks with transformers.

\section{Related work}
\label{sec:related}

Our work is based on the vision transformer (ViT)~\cite{dosovitskiy2020image}.
We briefly recall it and its follow-ups in this section.

\paragraph{Vision transformers.}
The pioneering work of ViT~\cite{dosovitskiy2020image} directly applies the transformer architecture to non-overlapping image patches for image classification.
It achieves impressive accuracy on image classification compared to convolutional neural networks, but ViT requires a large training data set (i.e. JFT-300M~\cite{sun2017revisiting}) to perform well.

Many recent studies have successfully improved its data efficiency.
DeiT~\cite{touvron2021training} used strong data augmentation and knowledge distillation to surpass EfficientNet~\cite{tan2019efficientnet} on ImageNet~\cite{deng2009imagenet}. BEiT~\cite{beit} and MAE~\cite{kaiming2021mae} successfully transferred the pre-training paradigm of natural language processing~\cite{devlin2018bert,radford2018improving,radford2019language} to vision, further improving the accuracy on ImageNet.
It should be noted that these methods intuitively create more training data, thus avoiding the transformer's appetite for data, without analyzing why vision transformers need more training data than convolutional neural networks.
At the same time, the existing data-efficient benchmark reports~\cite{zhang2021aggregating,hassani2021escaping,Chen_2021_visformer} are effective but lower than convnets. Therefore, we study the data-efficiency vision transformers from architecture.

\paragraph{Hybrid vision transformers.}
The hybrid architectures use convolution to introduce locality for transformer, which is used for acceleration in ViT~\cite{dosovitskiy2020image} because the computational complexity of the local structure is lower than that of the global counterpart.
More recently, it has evolved into an architecture that integrates local and global features to improve accuracy.
CoAtNet~\cite{dai2021coatnet} found that using convolution in the early stages of the network made model training more stable; Visformer~\cite{Chen_2021_visformer} explored the best mode of combining Self-Attention and ResNet; CCT~\cite{hassani2021escaping} verifies the data efficiency of hybrid architecture by introducing convolution in the shallow layer of the network.
There are various indications~\cite{li2021localvit,srinivas2021bottleneck,Yuan_2021_T2T} that introducing locality into the vision transformer can improve the stability of the model.

However, Cordonnier et al.~\cite{Cordonnier2020On} show that localized self-attention is similar to convolution, so the improvement of hybrid network accuracy is in contradiction with the need to construct the global receptive field network.
On the premise that Dosovitskiy et al.~\cite{dosovitskiy2020image} showed that the global receptive field network has advantages, we believe that the current vision transformer has shortcomings in learning local features.
Therefore, we consider introducing locality to assist the training of the global receptive field network.

\paragraph{Local vision transformers.}
Local vision transformers no longer introduce locality through convolution, but use self-attention in fixed or sliding windows.
Approaching successful convolutional neural networks is the main direction of research in this field, such as hierarchical structure, locality, and relativity.
For locality, fixed-window-based self-attention is usually used because it is easier to reduce computational complexity than sliding-window-based self-attention, and their final accuracy is similar.
This approach results in missing attention connections across windows. To make up for this defect, Swin~\cite{liu2021Swin} designed a window shift strategy to facilitate feature communication between windows, HeloNet~\cite{vaswani2021scaling} used halo to make the window larger, and NesT~\cite{zhang2021aggregating} simply aggregates feature maps.
They tend to report higher accuracy for their local networks (which may not be surprising), but such pure self-attention networks are more indicative of the original ViT's difficulty in learning local features than hybrid architectures.

Since we want to design a network with the global receptive field, we choose self-attention based on sliding windows which is convenient to implement. Therefore, the main problem we face is how to make the network initialize with a local receptive field, and make this local receptive field expand or shrink through training.
\section{Approach}
\label{sec:approach}

In this section, we first describe how to introduce locality on the self-attention module and make the receptive field learnable, then show how to initialize the receptive field of each self-attention head across the network.

\subsection{Self-Attention with Position}
\label{sec:sawithposition}

It is generally known that the self-attention module cannot learn the positional relation. ViT~\cite{dosovitskiy2020image} makes up for the lack of positional relation through position embedding, but it still cannot control the receptive field of each self-attention head which causes the network to pay too much attention to the overall situation. We control the receptive field by introducing attention bias $B$ to the self-attention module, so that the network can take into account the local features while learning the global features. 
Formally, we consider defining the self-attention with attention bias as:
\begin{align}
\mathrm{Attention}(Q,K,V) = \mathrm{Softmax}(\frac{QK^{T}}{\sqrt{d_k}} + B) V
\end{align}
where $Q$, $K$, $V$ are the input vectors of the layers considered, the $\mathrm{Softmax}$ function is applied over each row of the input matrix and the $\sqrt{d_k}$ term provides appropriate normalization. The $B$ can introduce positional relation for the self-attention module.
We next describe how to introduce locality and relativity via $B$, and explore its learnable possibilities.

\paragraph{Introducing locality.}
Our approach to introducing locality via $B$ is similar to masking, primarily suppressing distant attention rather than removing it, termed ``Focal Attention Bias".
The suppression strategy is simple: we only suppress attention outside the set window and do not process attention within the window.
Due to the settings, each self-attention head will have a fixed receptive field.

\begin{figure}[t]
  \centering
  \includegraphics[width=1\linewidth]{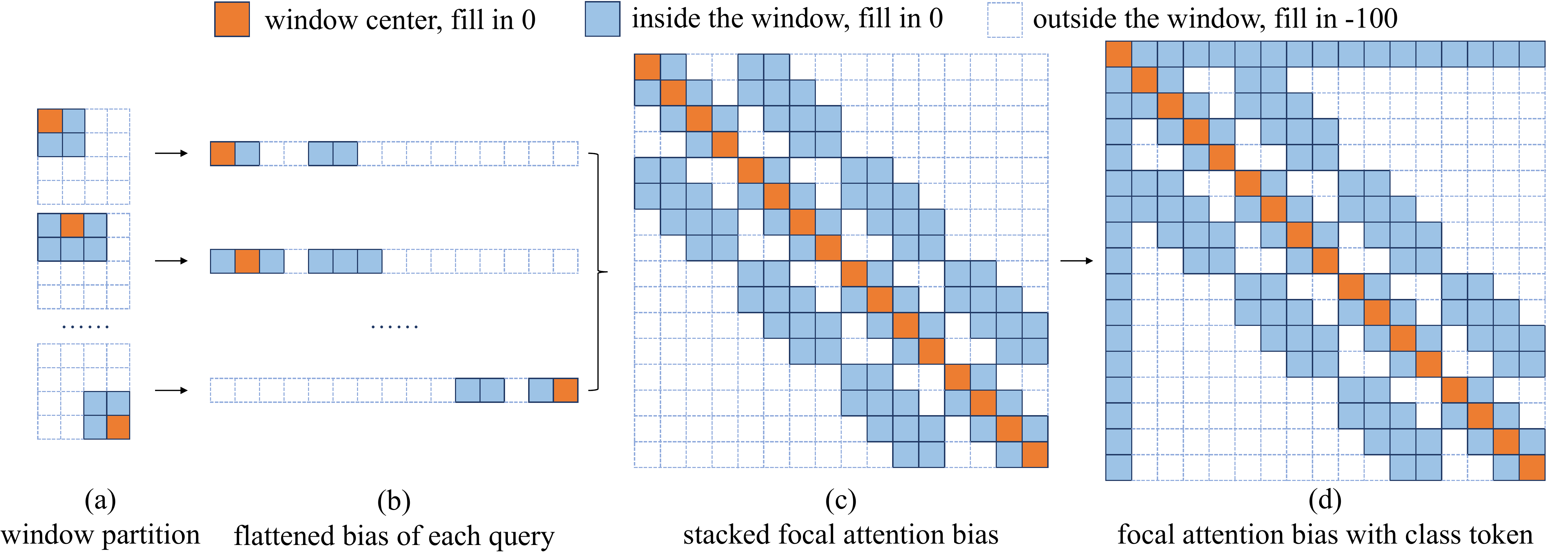}
  \caption{
  Illustration of focal attention bias for $3\times3$ window.
  The short-range attention is filled with $0$, and the long-range attention is filled with $-100$.
  In practice, we adopt multi-head self-attention with multi-scale receptive fields.
  }
  \label{fig:generate_prior}
\end{figure}

Specifically, since the product of $Q$ and $K$ yields an attention matrix of size $N \times N$, we set the focal attention bias $B$ to be the same size as $N\times N$.
We take the coordinates of each query token as the center and suppress the attention of key tokens outside the window.
Therefore, we need a parameter $w$ to represent the window size for each focal attention bias, and the larger window represents the larger receptive field.
If key token $j$ is outside the window centered on query token $i$, then $B_{ij}=-100$ (we verified this value in Table~\ref{table:diff_degree}), otherwise $B_{ij}=0$.
Figure~\ref{fig:generate_prior} presents the generation process of focal attention bias $P$ for $3\times3$ window.

\paragraph{Introducing relativity.}
Our focal attention bias approach has introduced relativity.
The relativity is introduced here to prevent over-fitting of the learnable focal attention bias, which forces self-attention to have the same attention bias at each location.

Following Swin~\cite{liu2021Swin}, the relative attention bias $B \in R^{M^2 \times M^2}$ along each axis lies in the range $[-M + 1, M - 1]$, we parameterize a smaller-sized matrix $\hat{B} \in R ^{(2M-1)\times(2M-1)}$, and values in $B$ are taken from $\hat{B}$.
Among them, $M^2$ is the number of image patches.
Since the relative attention bias is centered on the coordinates of the query token, we initialize $\hat{B}$ with a window centered on the coordinate $(0, 0)$.

The relativity we introduce is only for $B$ and does not affect $\frac{QK^{T}}{\sqrt{d_k}}$, so this approach is still learning the receptive field of the self-attention (implies locality assistance).
Since we want to show that this is an improvement over locality, most of our ablations do not use relative focal attention bias.

Our observation is consistent with Swin that relative focal attention bias can make the network perform better, as shown in Table~\ref{table:ab_or_re_pe}.
More importantly, we observe the receptive field of the self-attention head more clearly.

\begin{figure}[t]
  \centering
  \includegraphics[width=1\linewidth]{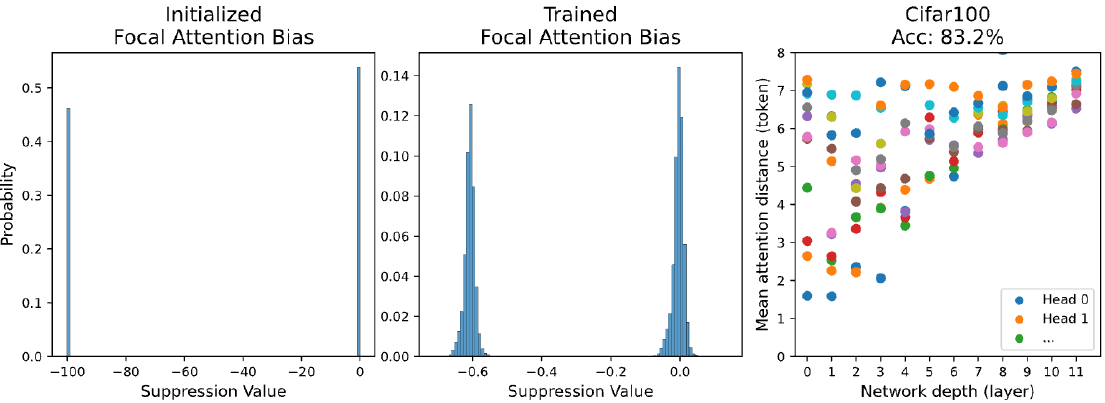}
  \caption{
  \textbf{Left \& Center:} Probability distribution of initialized and trained focal attention biases.
  When initialized, the suppression attention bias is padded with $-100$, and the non-suppression attention bias is padded with $0$.
  After training, the suppression values are all close to $0$, allowing the network to learn globally.
  \textbf{Right:} Mean attention distance of ViT-P Base trained from scratch on Cifar100.
  Compared with ViT, ViT-P has more local attention heads.
  In this experiment, the image width is $32$ pixels and the image patch is $2$ pixels.
  }
  \label{fig:trained_bias}
\end{figure}

\paragraph{Learnable focal attention bias.}
The easiest way is to directly make $B$ learnable, which already allows focal attention bias to learn its receptive field on a global scale.
The problem is the localization strategy we adopt will cause the gradient outside the window to be extremely small, so the attention bias $B$ can only shrink the receptive field and is difficult to enlarge.
Attention outside the window is reactivated by $\frac{QK^{T}}{\sqrt{d_k}}$ only in extreme cases.

Our solution is to gradually reduce the suppression of out-of-window attention with training.
This forces the network to learn features at different scales rather than just the global in early training, and adjust the best receptive field by itself in later training. We hypothesize that the network will not degenerate into global preferences if the local features are better.
Specifically, we introduce weight decay for $B$, which makes the weights automatically approach $0$ during training.

Following \cite{stephen1988wd,ily2019adamw}, weight decay is multiplying the parameter by a decay coefficient to make it closer to $0$ (from $-100$) when the gradient is updated, usually used to prevent over-fitting, but here used for de-suppression.
Formally, the weights $\theta$ decay exponentially as:
\begin{align}
\theta_{t+1} = (1 - \lambda) \theta_{t} - \alpha \triangledown f_{t}(\theta_{t})
\end{align}
where $\lambda$ defines the rate of the weight decay per step and $\triangledown f_{t}(\theta_{t})$ is the t-th batch gradient to be multiplied by a learning rate $\alpha$. 
Figure~\ref{fig:trained_bias} shows that using weight decay for focal attention bias is consistent with our purpose.

\subsection{Network Architectures}
\label{sec:network_architectures}

\paragraph{Plain network.}
Our plain baseline is inherited from ViT~\cite{dosovitskiy2020image} and DeiT~\cite{touvron2021training}, using the uniform-scale network, the class token, and the absolute position embedding.
Based on the network, we insert focal attention bias $B$ in each self-attention head and convert the network to the corresponding version with locality and relativity (which is used to assist locality learning).

\paragraph{Overall architecture.}
Considering that we are doing attention clipping on the global attention network, the correct clipping method is very important.
To be clear, we should not abandon long-range attention, as Dosovitskiy et al.~\cite{dosovitskiy2020image} have demonstrated long-range attention to be effective on large data sets. Therefore, we only consider the fusion architecture of short-range and long-range attention. For convenience, we set the receptive field of attention as a linear scaling from the local sliding window to the global. We simply refer to this as ``Multi-scale Receptive Field Attention".

\begin{figure}[t]
  \centering
  \includegraphics[width=1\linewidth]{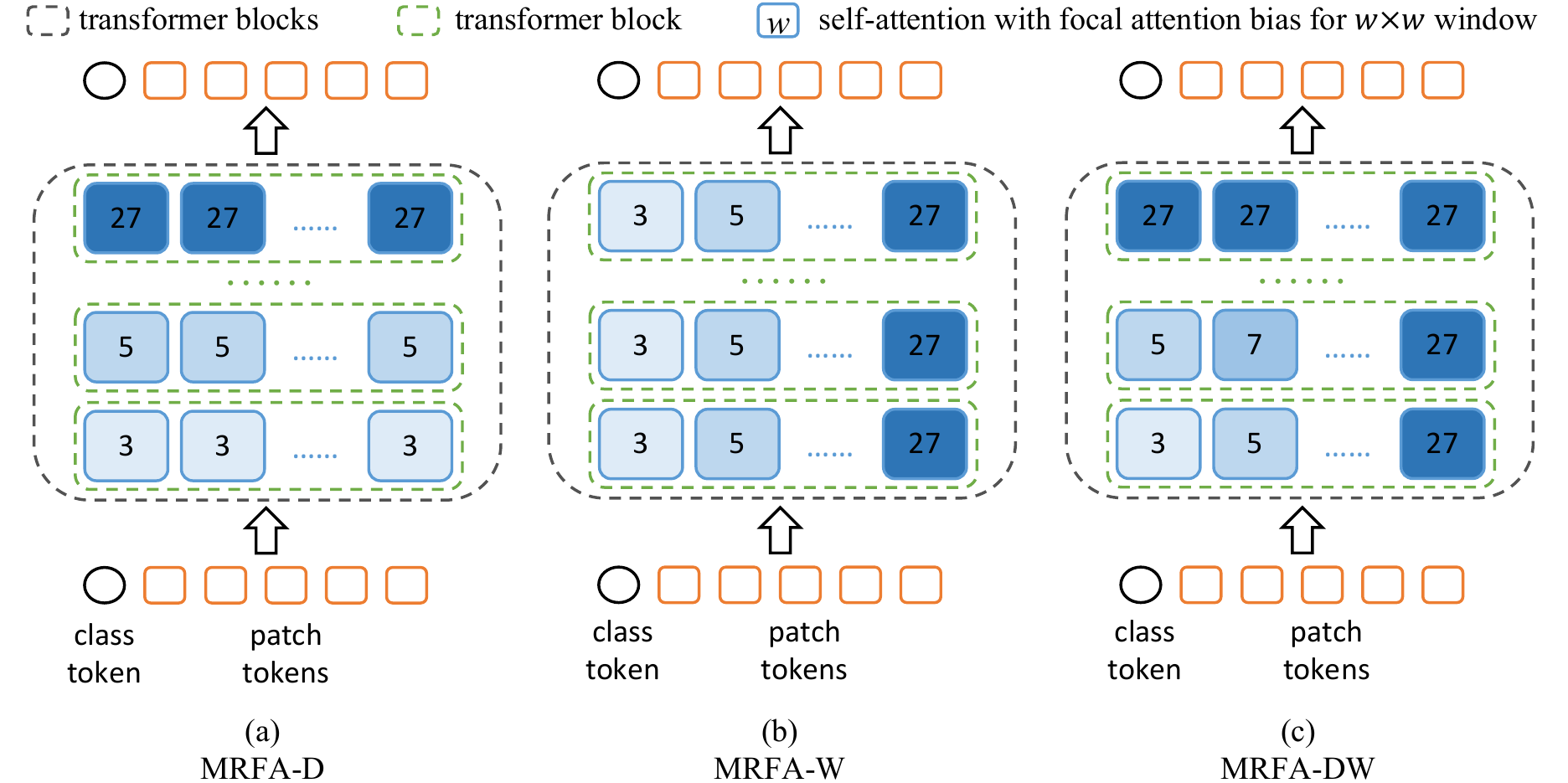}
  \caption{
    Three initialization modes of ViT-P.
    These three modes can roughly generalize the distribution patterns of the vision transformers' local and global attention heads.
    We adopted the best performing MRFA-W after ablation.
    Furthermore, we find that de-suppression during training is beneficial, indicating that multi-scale receptive field attention can assist the training of global receptive field networks.
  }
  \label{fig:diff_mode}
\end{figure}

\paragraph{Network with multi-scale receptive fields attention.}
We summarize three modes for multi-scale receptive field attention (MRFA), namely multi-scale receptive field attention in depth (MRFA-D), width (MRFA-W), and both width and depth (MRFA-DW).
Similar counterparts can be found in existing methods~\cite{liu2021Swin,d2021convit} for each mode, illustrating how important it is to combine short-range and long-range attention.
And we experimentally verify the effectiveness of these three modes in Table~\ref{table:diff_mode}, where the MRFA-W is the best.
Considering that we still want to learn spatial attention, the minimum window size is $3 \times 3$. The following is the detail of three modes:

\textbf{MRFA-D} expands the receptive field of attention with increasing depth, as shown in Figure~\ref{fig:diff_mode}a. Specifically, all the heads in the first layer of the network have the receptive field with $3 \times 3$ local window, while the heads in the last layer of the network have the global receptive field, and the layers between them have an equal step size to increase the receptive field. 

\textbf{MRFA-W} has multi-scale receptive field attention on each multi-head attention layer, as shown in Figure~\ref{fig:diff_mode}b. Specifically, each attention head at the layer increases the receptive field from the $3 \times 3$ local window to the global using the equal step.

\textbf{MRFA-DW} is the combination of MRFA-D and MRFA-W, which has multi-scale receptive fields in both depth and width, as shown in Figure~\ref{fig:diff_mode}c. Specifically, the network has multi-scale receptive fields in width, and the minimum receptive field of attention heads increases with depth.

\section{Experiments}
\label{sec:experiments}

\paragraph{Setting.}
We refer to previous work~\cite{touvron2021training} to generate our models, namely Tiny, Small and Base.
Many recent ViT-based methods refer to the training technique of DeiT~\cite{touvron2021training}, and we note that MAE~\cite{kaiming2021mae} proposes a better training technique based on it.
We trained with the setting of MAE on ImageNet but slightly modified it on Cifar.
On Cifar, we use a wd of 0.3, a long warmup of 20 and $\beta_2=0.95$ following \cite{kaiming2021mae}, and adjust the learning rate to 5e-4$ \times \frac{\mathrm{batch size}}{256}$ and batch sizes to $64$ (base), $128$ (small), and $256$ (tiny).
Based on this setup, we can train ViT-P on 1 GPU.

\paragraph{Dataset: Cifar.}
Since our goal is to efficiently train vision transformers on small data sets, we did not use the more popular ImageNet with 1 million training data for ablation, but Cifar with only 50,000 training data.

\paragraph{Baseline: ViT-Base.}
We use ViT-Base~\cite{touvron2021training} as the baseline in our ablation study. ViT-B of the same depth and width can better illustrate our ablation. Additionally, we note that ViT does not yet have standard training hyper-parameters on Cifar and a good recipe with strong regularization is needed to be established first. Here the training uses $32 \times 32$ input image and $2 \times 2$ image patch embedding. The following is the comparison of our implementation with the previous work:
\setlength{\tabcolsep}{4pt}
\begin{center}\vspace{-.6em}
\label{table:baseline}
\begin{tabular}{@{}c|cc@{}}
\hline\noalign{\smallskip}
ViT-B/2 & Cifar10 (\%) & Cifar100 (\%) \\
\noalign{\smallskip}
\hline\noalign{\smallskip}
prev best~\cite{zhang2021aggregating} & 92.41 & 70.49 \\
our impl.                             & 94.02 & 76.48 \\
\hline
\end{tabular}
\end{center}\vspace{-.6em}
\setlength{\tabcolsep}{1.4pt}

\subsection{Main Properties}
\label{sec:ablation}

\paragraph{Different modes of MRFA.}
Table~\ref{table:diff_mode} shows the effect of different local modes.
It is important to introduce locality correctly, otherwise, the network needs to learn on its own.
ViT lets the network learn local experience by itself, and we try to introduce this through initialization.
The three local modes we consider can cover all potential modes with rough but simple initialization settings, and we hypothesize that the network can be optimized by training to obtain a better local experience.
This design can yield up to 4\% improvement (or 7\% if further relativity is introduced) when using MRFA-W mode on Cifar100.
Furthermore, we observe that competitive results can be obtained without using weight decay for focal attention bias $B$ to make suppressed attention difficult to reactivate.

Interestingly, even if our model uses a fixed attention bias as a subset of ViT~\cite{dosovitskiy2020image}, the results are better than the original. This shows that ViT sets excessive long-range attention, which makes it difficult to learn local features.
Unsurprisingly, it is expected to improve with the introduction of learnable attention bias, showing potential benefits of global architectures that require large amounts of data for training or introduce similar experiences.

Overall, Multi-scale Receptive Field Attention in Width (MRFA-W) has the best performance, which we believe is related to the fusion of different receptive field features.
When using learnable focal attention bias with weight decay, the mean attention distance of our model trained on Cifar is similar to that of ViT trained on a larger data set, indicating that we successfully fed experience similar to learning on large data sets into the network.
\setlength{\tabcolsep}{4pt}
\begin{table}[t]
\begin{center}
\caption{
    Compare the accuracy of ViT-P with different modes on Cifar.
    We conduct three sets of experiments for each mode, namely fixed focal attention bias, learnable focal attention bias, and learnable focal attention bias with weight decay.
    The suppression value is uniformly -100, and relativity is not introduced.
}
\label{table:diff_mode}
\begin{tabular}{@{}ccc|cc@{}}
\toprule
Mode                     & Learnable Bias & Bias with Decay  & Cifar10 (\%)    & Cifar100 (\%)  \\ \midrule
\multirow{3}{*}{MRFA-DW} & \xmark             & \xmark               & 96.07           & 78.38          \\
                         & \cmark             & \xmark               & 96.28           & 79.02          \\
                         & \cmark             & \cmark               & 96.32           & 78.84          \\ \midrule
\multirow{3}{*}{MRFA-D}  & \xmark             & \xmark               & 95.71           & 77.79          \\
                         & \cmark             & \xmark               & 95.77           & 78.12          \\
                         & \cmark             & \cmark               & 95.83           & 78.43          \\ \midrule
\multirow{3}{*}{MRFA-W}  & \xmark             & \xmark               & 96.54           & 79.48          \\
                         & \cmark             & \xmark               & 96.65           & 80.44          \\
                         & \cmark             & \cmark               & \textbf{96.96}  & \textbf{80.62} \\ \bottomrule
\end{tabular}
\end{center}
\end{table}
\setlength{\tabcolsep}{1.4pt}

\paragraph{Suppression value of focal attention bias.}
The suppression value is an important parameter of focal attention bias, which directly affects the reactivation difficulty of the suppressed attention.
And the strength of suppression using learnable focal attention bias with weight decay is negatively related to the number of steps in training. 
Table~\ref{table:diff_degree} ablated suppression values using the learnable focal attention bias with weight decay for MRFA-W.

If the suppression value is set to $-100$, we treat the attention as being removed. In this case, our design approximates mask-based self-attention when initialized.
Masking is not used directly because we want the suppressed attention to gradually reactivate with training, and a smaller suppression value means earlier reactivation of attention.
As the degree of suppression decreases, the network gradually degenerates to ViT.
In our settings, the suppression value of $-50$ worked best on Cifar10 and $-100$ worked best on Cifar100. And we finally chose $-100$ as the suppression value.
Our results indicate that premature degradation (to the original ViT) degrades performance on small data sets.
Also, slow or insufficient degradation can negatively impact accuracy.

Moreover, we try to set the suppression value to $0$ to verify the effect of introducing a random initialization attention bias.
It turns out that this randomly initialized design is helpful for the network when attention bias $B$ is introduced, and the further introduction of multi-scale receptive field attention can improve the accuracy of the network again.

\setlength{\tabcolsep}{4pt}
\begin{table}[t]
\begin{center}
\caption{
    Accuracy of ViT-P with different suppression values using MRFA-W without relativity.
    A larger suppression value means the network is harder to activate attention and suppressed for longer during training.
}
\label{table:diff_degree}
\begin{tabular}{@{}c|cc@{}}
\toprule
Suppression Value & Cifar10 (\%)   & Cifar100 (\%)  \\ \midrule
-100              & 96.96          & \textbf{80.62} \\
-50               & \textbf{97.06} & 80.49          \\
-10               & 96.99          & 80.54          \\
-5                & 96.91          & 79.95          \\
-1                & 96.59          & 79.09          \\ 
0                 & 96.47          & 78.56          \\ \bottomrule
\end{tabular}
\end{center}
\end{table}
\setlength{\tabcolsep}{1.4pt}
\setlength{\tabcolsep}{4pt}
\begin{table}[t]
\begin{center}
\caption{
    Absolute focal attention bias vs. Relative focal attention bias. The suppression value is -100. Relativity is beneficial with and without weight decay.
}
\label{table:ab_or_re_pe}
\begin{tabular}{@{}cc|cc@{}}
\toprule
Bias Method               & Decay  & Cifar10 (\%)   & Cifar100 (\%)  \\ \midrule
\multirow{2}{*}{Absolute} & \xmark & 96.65          & 80.44          \\
                          & \cmark & 96.96          & 80.62          \\ \midrule
\multirow{2}{*}{Relative} & \xmark & 97.02          & 82.85          \\
                          & \cmark & \textbf{97.20} & \textbf{83.16} \\ \bottomrule
\end{tabular}
\end{center}
\end{table}
\setlength{\tabcolsep}{1.4pt}

\paragraph{Relative position.}
We compare the effects of relativity in Table~\ref{table:ab_or_re_pe}.
Our results so far are based on the absolute focal attention bias.
Using relative focal attention bias can improve accuracy.
This relativity prevents the attention bias from over-fitting.
We hypothesize that different windows have consistent preferences for input images concerning location. If this preference is not introduced, the network will learn by itself. Our experiments show that this experience may be difficult to learn when data is scarce.

We also compare relative attention bias schemes without weight decay.
The results show that vision transformers can perform well even without activating suppressed attention, although they cannot outperform the ones with weight decay.

\setlength{\tabcolsep}{4pt}
\begin{table}[t]
\begin{center}
\caption{
    Results on Cifar.
    All methods are trained from scratch using Cifar data set and take images of size $32\times32$.
    We use the ViT-like notation~\cite{dosovitskiy2020image} for model size and patch size: for instance, Our-B/2 denotes the ``Base'' variant with $2\times2$ input patch size.
    ${^\dagger}$: Our implementation.
}
\label{table:compare_cifar}
\begin{tabular}{@{}cc|r|cc@{}}
\toprule
Architecture                                             & Method              & Params & Cifar10 (\%)   & Cifar100 (\%)   \\ \midrule
\multirow{2}{*}{Convnet}                                 & Pyramid-164-48      & 1.7M    & 95.97          & 80.70          \\ \cmidrule(l){2-5} 
                                                         & WRN28-10            & 36.5M   & 95.83          & 80.75          \\ \midrule
\multirow{9}{*}{\makecell[c]{Multi-stage\\Transformer}}  & Swin-T/1            & 27.5M   & 94.46          & 78.07          \\
                                                         & Swin-S/1            & 48.8M   & 94.17          & 77.01          \\
                                                         & Swin-B/1            & 86.7M   & 94.55          & 78.45          \\ \cmidrule(l){2-5}
                                                         & NesT-T/2            & -       & 94.53          & 77.13          \\
                                                         & NesT-S/2            & -       & 95.87          & 79.06          \\
                                                         & NesT-B/2            & -       & 96.09          & 79.89          \\
                                                         & NesT-T/1            & 6.2M    & 96.04          & 78.69          \\
                                                         & NesT-S/1            & 23.4M   & 96.97          & 81.70          \\
                                                         & NesT-B/1            & 90.1M   & 97.20          & 82.56          \\ \midrule
\multirow{8}{*}{\makecell[c]{Single-stage\\Transformer}} & DeiT-T/2            & 5.3M    & 88.39          & 67.52          \\
                                                         & DeiT-S/2            & 21.3M   & 92.44          & 69.78          \\
                                                         & DeiT-B/2            & 85.1M   & 92.41          & 70.49          \\
                                                         & ViT-B/2${^\dagger}$ & 85.1M   & 94.02          & 76.48          \\ \cmidrule(l){2-5}
                                                         & CCT-7/3$\times$1    & 3.7M    & 94.72          & 76.67          \\ \cmidrule(l){2-5}
                                                         & Our-T/2             & 5.4M    & 95.91          & 81.36          \\
                                                         & Our-S/2             & 21.4M   & 96.98          & 81.93          \\
                                                         & Our-B/2             & 85.3M   & \textbf{97.20} & \textbf{83.16} \\ \bottomrule
\end{tabular}
\end{center}
\end{table}
\setlength{\tabcolsep}{1.4pt}

\subsection{Comparisons with Previous Results}

\paragraph{Cifar.}
We compare recent approaches trained from scratch on Cifar data sets~\cite{krizhevsky2009learning} in Table~\ref{table:compare_cifar}, to investigate data efficiency.
At present, methods that can improve the data efficiency of transformers have emerged, which shows the potential of transformer-based methods on small data sets.

Our ViT-P is easily scalable using a single-scale network and shows significant improvement with equivalent parameters.
We obtained 97.20\% (Cifar10) and 83.16\% (Cifar100) accuracy using ViT-P Base.
Among all transformer-based methods using only Cifar data, the previous best accuracy is 82.56\%~\cite{zhang2021aggregating} on Cifar100, based on advanced networks.
Our result is based on ViT~\cite{dosovitskiy2020image}, and we expect advanced networks will perform better.

Comparing with the follow-ups~\cite{wang2021pvt,liu2021Swin,zhang2021aggregating}, our ViT-P is more accurate while being simpler.
They typically use the multi-stage network.
We drop it because we don't think it is the key to data efficiency, but this structure could lead to further improvements.
More importantly, we show that even with the single-stage network, the accuracy of our network is comparable to current multi-stage networks.

In addition, Swin and NesT also introduce locality, and the comparison shows the strong advantage of using multi-scale receptive field attention over the width of the network.

\setlength{\tabcolsep}{4pt}
\begin{table}[t]
\begin{center}
\caption{
    Results on ImageNet.
    All results are on an image size of 224.
    Here our ViT-P uses the same training hyperparameters as MAE~\cite{kaiming2021mae}.
    ${^\dagger}$: MAE implementation.
}
\label{table:compare_imagenet}
\begin{tabular}{@{}cc|r|c@{}}
\toprule
Architecture                                             & Method               & Params & ImageNet (\%) \\ \midrule
\multirow{2}{*}{Convnet}                                 & ResNet-50            & 25M    & 76.2          \\ \cmidrule(l){2-4} 
                                                         & RegNetY-4G           & 21M    & 80.0          \\ \midrule
\multirow{6}{*}{\makecell[c]{Multi-stage\\Transformer}}  & Swin-T/4             & 29M    & 81.3          \\
                                                         & Swin-S/4             & 50M    & 83.0          \\
                                                         & Swin-B/4             & 88M    & 83.5          \\ \cmidrule(l){2-4} 
                                                         & NesT-T/4             & 17M    & 81.5          \\
                                                         & NesT-S/4             & 38M    & 83.3          \\
                                                         & NesT-B/4             & 68M    & \textbf{83.8} \\ \midrule
\multirow{7}{*}{\makecell[c]{Single-stage\\Transformer}} & DeiT-T/16           & 6M     & 72.2          \\
                                                         & DeiT-S/16            & 22M    & 79.8          \\
                                                         & DeiT-B/16            & 87M    & 81.8          \\ \cmidrule(l){2-4}
                                                         & ViT-B/16${^\dagger}$ & 87M    & 82.3          \\ \cmidrule(l){2-4}
                                                         & Our-T/16             & 6M     & 72.4          \\
                                                         & Our-S/16             & 22M    & 80.5          \\
                                                         & Our-B/16             & 87M    & 82.1          \\ \bottomrule
\end{tabular}
\end{center}
\end{table}
\setlength{\tabcolsep}{1.4pt}

\paragraph{ImageNet.}
We test ViT-P on standard ImageNet 2012 benchmarks~\cite{deng2009imagenet} with commonly used 300 epoch training in Table~\ref{table:compare_imagenet} to show that our approach does not lose accuracy on larger data sets. The input size is $224 \times 224$ and no extra pre-training data is used.

Compared to ViT-P, Swin~\cite{liu2021Swin} and NesT~\cite{zhang2021aggregating} use different configurations, which allows the network to achieve higher accuracy with less parameter amount.
The main difference is that the depth and the number of heads are increased and the dimension of each self-attention head is reduced (from $64$ to $32$), which means a more complex structure compared to linear layers.
Moreover, the multi-stage network may have potential benefits.
The purpose of this paper is not to enjoy these benefits, but to demonstrate by comparison that ViT can improve data efficiency by introducing locality, thus not changing the model scale.

ViT-P can be regarded as a subset network of ViT~\cite{dosovitskiy2020image}, and its accuracy is comparable on large data sets after suppressing partial attention during initialization.
This observation suggests that vision transformers has excessive spatial attention.
And MRFA-W, as a suppression method with comparable performance, enables the vision transformer to be applied to visual data sets of various sizes.
 
\section{Conclusion}
\label{sec:conclusion}

We illustrate the superiority of a ViT-like global architecture, and experimentally illustrate the limitations of ViT on local feature learning.
Introducing locality enables global networks to compete with convolutional neural networks on small data sets.
On this basis, we can easily and confidently apply self-supervised learning methods~\cite{kaiming2021mae,beit,radford2018improving,devlin2018bert,radford2019language}, as these methods often increase training data diversity, and predicting or comparing image details allows the network to learn local information autonomously.
We hope that this idea of introducing locality to aid global network learning inspires future works.

We provide an open-source implementation of our method. It is available at
\url{https://github.com/freder-chen/vitp}.


{\small
\bibliographystyle{ieee_fullname}
\bibliography{main}

\begin{thebibliography}{10}\itemsep=-1pt

\bibitem{beit}
Hangbo Bao, Li Dong, and Furu Wei.
\newblock {BEiT}: {BERT} pre-training of image transformers.
\newblock 2021.

\bibitem{carion2020detr}
Nicolas Carion, Francisco Massa, Gabriel Synnaeve, Nicolas Usunier, Alexander
  Kirillov, and Sergey Zagoruyko.
\newblock End-to-end object detection with transformers.
\newblock In {\em European Conference on Computer Vision}, 2020.

\bibitem{Chen_2021_visformer}
Zhengsu Chen, Lingxi Xie, Jianwei Niu, Xuefeng Liu, Longhui Wei, and Qi Tian.
\newblock Visformer: The vision-friendly transformer.
\newblock In {\em International Conference on Computer Vision}, 2021.

\bibitem{Cordonnier2020On}
Jean-Baptiste Cordonnier, Andreas Loukas, and Martin Jaggi.
\newblock On the relationship between self-attention and convolutional layers.
\newblock In {\em International Conference on Learning Representations}, 2020.

\bibitem{dai2021coatnet}
Zihang Dai, Hanxiao Liu, Quoc~V. Le, and Mingxing Tan.
\newblock {CoAtNet}: Marrying convolution and attention for all data sizes.
\newblock 2021.

\bibitem{d2021convit}
St{\'e}phane d'Ascoli, Hugo Touvron, Matthew~L. Leavitt, Ari~S. Morcos, Giulio
  Biroli, and Levent Sagun.
\newblock {ConViT}: Improving vision transformers with soft convolutional
  inductive biases.
\newblock In {\em International Conference on Machine Learning}, 2021.

\bibitem{deng2009imagenet}
Jia Deng, Wei Dong, Richard Socher, Li-Jia Li, Kai Li, and Li Fei-Fei.
\newblock {ImageNet}: A large-scale hierarchical image database.
\newblock In {\em Computer Vision and Pattern Recognition}, 2009.

\bibitem{devlin2018bert}
Jacob Devlin, Ming-Wei Chang, Kenton Lee, and Kristina Toutanova.
\newblock {BERT}: Pre-training of deep bidirectional transformers for language
  understanding.
\newblock In {\em North American Chapter of the Association for Computational
  Linguistics}, 2018.

\bibitem{dosovitskiy2020image}
Alexey Dosovitskiy, Lucas Beyer, Alexander Kolesnikov, Dirk Weissenborn,
  Xiaohua Zhai, Thomas Unterthiner, Mostafa Dehghani, Matthias Minderer, Georg
  Heigold, Sylvain Gelly, Jakob Uszkoreit, and Neil Houlsby.
\newblock An image is worth 16x16 words: Transformers for image recognition at
  scale.
\newblock In {\em International Conference on Learning Representations}, 2021.

\bibitem{stephen1988wd}
Stephen~Jos{\'e} Hanson and Lorien Pratt.
\newblock Comparing biases for minimal network construction with
  back-propagation.
\newblock In {\em Neural Information Processing Systems}, 1988.

\bibitem{hassani2021escaping}
Ali Hassani, Steven Walton, Nikhil Shah, Abulikemu Abuduweili, Jiachen Li, and
  Humphrey Shi.
\newblock Escaping the big data paradigm with compact transformers.
\newblock {\em arXiv}, 2021.

\bibitem{kaiming2021mae}
Kaiming He, Xinlei Chen, Saining Xie, Yanghao Li, Piotr Doll\'ar, and Ross
  Girshick.
\newblock Masked autoencoders are scalable vision learners.
\newblock 2021.

\bibitem{he2016deep}
Kaiming He, Xiangyu Zhang, Shaoqing Ren, and Jian Sun.
\newblock Deep residual learning for image recognition.
\newblock In {\em Computer Vision and Pattern Recognition}, 2016.

\bibitem{krizhevsky2009learning}
Alex Krizhevsky.
\newblock Learning multiple layers of features from tiny images.
\newblock 2009.

\bibitem{NIPS2012_c399862d}
Alex Krizhevsky, Ilya Sutskever, and Geoffrey~E. Hinton.
\newblock Imagenet classification with deep convolutional neural networks.
\newblock In {\em Neural Information Processing Systems}, 2012.

\bibitem{li2021localvit}
Yawei Li, Kai Zhang, Jiezhang Cao, Radu Timofte, and Luc~Van Gool.
\newblock {LocalViT}: Bringing locality to vision transformers.
\newblock {\em arXiv}, 2021.

\bibitem{liu2021Swin}
Ze Liu, Yutong Lin, Yue Cao, Han Hu, Yixuan Wei, Zheng Zhang, Stephen Lin, and
  Baining Guo.
\newblock {Swin Transformer}: Hierarchical vision transformer using shifted
  windows.
\newblock {\em International Conference on Computer Vision}, 2021.

\bibitem{ily2019adamw}
Ilya Loshchilov and Frank Hutter.
\newblock Decoupled weight decay regularization.
\newblock In {\em International Conference on Learning Representations}, 2019.

\bibitem{radford2018improving}
Alec Radford, Karthik Narasimhan, Tim Salimans, and Ilya Sutskever.
\newblock Improving language understanding by generative pre-training.
\newblock {\em arXiv}, 2018.

\bibitem{radford2019language}
Alec Radford, Jeffrey Wu, Rewon Child, David Luan, Dario Amodei, Ilya
  Sutskever, et~al.
\newblock Language models are unsupervised multitask learners.
\newblock {\em OpenAI blog}, 2019.

\bibitem{srinivas2021bottleneck}
Aravind Srinivas, Tsung-Yi Lin, Niki Parmar, Jonathon Shlens, Pieter Abbeel,
  and Ashish Vaswani.
\newblock Bottleneck transformers for visual recognition.
\newblock {\em arXiv}, 2021.

\bibitem{sun2017revisiting}
Chen Sun, Abhinav Shrivastava, Saurabh Singh, and Abhinav Gupta.
\newblock Revisiting unreasonable effectiveness of data in deep learning era.
\newblock In {\em International Conference on Computer Vision}, 2017.

\bibitem{tan2019efficientnet}
Mingxing Tan and Quoc~V. Le.
\newblock {EfficientNet}: Rethinking model scaling for convolutional neural
  networks.
\newblock In {\em International Conference on Machine Learning}, 2019.

\bibitem{touvron2021training}
Hugo Touvron, Matthieu Cord, Matthijs Douze, Francisco Massa, Alexandre
  Sablayrolles, and Herv{\'e} J{\'e}gou.
\newblock Training data-efficient image transformers \& distillation through
  attention.
\newblock In {\em International Conference on Machine Learning}, 2021.

\bibitem{vaswani2021scaling}
Ashish Vaswani, Prajit Ramachandran, Aravind Srinivas, Niki Parmar, Blake~A.
  Hechtman, and Jonathon Shlens.
\newblock Scaling local self-attention for parameter efficient visual
  backbones.
\newblock In {\em Computer Vision and Pattern Recognition}, 2021.

\bibitem{wang2021pvt}
Wenhai Wang, Enze Xie, Xiang Li, Deng-Ping Fan, Kaitao Song, Ding Liang, Tong
  Lu, Ping Luo, and Ling Shao.
\newblock Pyramid vision transformer: A versatile backbone for dense prediction
  without convolutions.
\newblock In {\em International Conference on Computer Vision}, 2021.

\bibitem{Yuan_2021_T2T}
Li Yuan, Yunpeng Chen, Tao Wang, Weihao Yu, Yujun Shi, Zi-Hang Jiang,
  Francis~E.H. Tay, Jiashi Feng, and Shuicheng Yan.
\newblock {Tokens-to-Token ViT}: Training vision transformers from scratch on
  imagenet.
\newblock In {\em International Conference on Computer Vision}, 2021.

\bibitem{zhang2021aggregating}
Zizhao Zhang, Han Zhang, Long Zhao, Ting Chen, , Sercan~Ö. Arık, and Tomas
  Pfister.
\newblock {Nested Hierarchical Transformer}: Towards accurate, data-efficient
  and interpretable visual understanding.
\newblock In {\em AAAI Conference on Artificial Intelligence}, 2022.

\end{thebibliography}
}

\end{document}